\documentclass{article} 
\usepackage{iclr2023_conference_tinypaper,times}

\usepackage{amsmath,amsfonts,bm}

\def\eqref#1{equation~\ref{#1}}

\def\1{\bm{1}}

\DeclareMathAlphabet{\mathsfit}{\encodingdefault}{\sfdefault}{m}{sl}
\SetMathAlphabet{\mathsfit}{bold}{\encodingdefault}{\sfdefault}{bx}{n}

\usepackage{hyperref}
\usepackage{url}
\usepackage{dirtytalk}
\usepackage{amsmath}
\usepackage{float}
\usepackage{algpseudocode}
\usepackage{hyperref}
\usepackage{url}
 \usepackage{algorithm}
\usepackage{dirtytalk}
\usepackage{siunitx}
\usepackage{mathtools}
\usepackage{multirow}
\usepackage{xr}
\usepackage{placeins}
\usepackage{siunitx}
\usepackage{mathtools}
\usepackage{float}
\usepackage{multirow}
\usepackage{xr}
\usepackage{authblk}
\title{Causal Covariate Shift Correction using Fisher information penalty}

\author[1,2]{Behraj Khan}
\author[3]{Behroz Mirza}
\author[1]{Tahir Syed}
\affil[1]{Institute of Business Administration Karachi, Pakistan \texttt{\{behrajkhan,tqsyed\}@iba.edu.pk} }
\affil[2]{National University of Computer and Emerging Sciences, Pakistan }
\affil[3]{Habib University, Karachi,  Pakistan \texttt{\{behroz.mirza\}@sse.habib.edu.pk} }

\iclrfinalcopy 
\begin{document}
\maketitle
\vspace{-0.5 cm}
\begin{abstract}
 Evolving feature densities across batches of training data bias cross-validation, making model selection and assessment unreliable (\cite{sugiyama2012machine}). This work takes a distributed density estimation angle to the training setting where data are temporally distributed. \textit{Causal Covariate Shift Correction ($C^{3}$)}, accumulates knowledge about the data density of a training batch using Fisher Information, and using it to penalize the loss in all subsequent batches. The penalty improves accuracy by $12.9\%$ over the full-dataset baseline, by $20.3\%$ accuracy at maximum in batchwise and  $5.9\%$  at minimum in foldwise benchmarks. 

\end{abstract}

\section{Background}

In learning systems over big data, the training dataset may not be completely available at the same time and place. Therefore, such a real-world setting fails the assumption of the classical machine learning model of independent and identically distributed (iid) source and target data.
Such problems are characterized as distribution shift \cite{quinonero2008dataset}, most commonnly as  \textit{covarite shift} \cite{cortes2008sample,sugiyama2007direct, moreno2012study, vigneron2010adaptive}, where training and test feature distributions differ. In continual or streaming applications, we may encounter what we term \textit{causal covariate shift}, causality being defined as the immutability of the batching sequence \cite{waveNet}.  

\section{Method development} 

Measuring the amount of shift between distribution of different batches requires the use of a metric on the space of probability distributions. The natural choice is the relative entropy / Kullback-Leibler divergence \cite{kullback1951information} as a quasi-metric because of its theoretical proximity with the cross entropy network loss, and because the ordering among batches imposes a natural direction on the computation.   The usual mean-field formulation of the KL-divergence uses a Gaussian  $q(\theta)$ distribution, parametrised with the covariance matrix of the network parameters. This term involves the Hessian of the derivatives of the parameters, and the high-dimensionality of the latter renders its computation intractable.  Recent inference literature \cite{pascanu2013revisiting} suggests an approximation by the Fisher Information Matrix (FIM), a quantity that  can be derived using the variance and expected value of the function of interest  \cite{nishiyama2019new}. %

   
Let us consider a model with parameters $\theta$ and a likelihood function \(p(X \mid \theta)\), where  \(X\) is observed data. The estimate of true parameter  $\theta$ can be found by using estimator $\hat{\theta}$. The Fisher information \(I(\theta)\) can be defined as the expected value of the negative hessian of  the log-likelihood function \(I(\theta) = \mathbb{E}\left[- \frac{\partial^2 \log p(X\mid\theta)}{\partial\theta\partial\theta^T}\right]\).  The Cram\'er-Rao Lower Bound (CRLB) states that for any unbiased estimator $\hat{\theta}$, the covariance matrix \(V(\hat{\theta})\) satisfies the matrix inequality property: \(
 V(\hat{\theta}) \succeq I^{-1}(\theta)\).

Now let us assume \(q(\hat{\theta}\)) as Gaussian distribution function to be   estimated around parameter $\theta$ mean and variance-covariance matrix \(V(\hat{\theta})\) in such manner that:
\(
    q(\hat{\theta}) \approx \mathcal{N}(\theta, V(\hat{\theta}))  
\). 
The $D_{KL}$ for source to target distribution is defined by $D_{KL}(P \,||\, Q) = \sum_y P(y|x) \log \left( \frac{P(y|x)}{Q(y|x)} \right)$, with $P(y|x)$ as the classifier output, and  $Q(y|x)$ as Normal with means $\mu_{P}$ and $\mu_{Q}$ and variances $\sigma^2_{P}$ and $\sigma^2_{Q}$. 

The $ D_{KL}$ can be approximated as:
\begin{equation}
  D_{KL}(p(\theta) \parallel q(\hat{\theta})) \approx \int p(\theta) \log\left( \frac{p(\theta)}{\mathcal{N}(\theta, V(\hat{\theta}))} \right) d\theta   
\end{equation}

Our method approximates the $V(\hat{\theta})$ by $I({\theta})$, with detail in  appendix A: 

 \begin{equation}
 I(\theta) = -\mathbb{E}\left[\frac{\partial^2 \log P(X;\theta)}{\partial \theta^2}\right]   
\end{equation}
An important result \cite{courtade2016monotonicity} shows that Fisher information remains strictly upper bounded by entropy, such as the increasing frontiers in \cite{vigneron2010adaptive}. We propose a massive decrease in computation by using the FIM instead of the entire divergence term. The formulation follows the familiar Tikhonov mechanism of weighted penalty addition. 
 \begin{equation}  
 \mathcal{L}(x, y; \theta) = - \int P(y(x)) \log(P(y|x; \theta))d\theta  - \lambda \times \int \frac{\partial^2 \log p(X\mid\theta)}{\partial\theta\partial\theta^T}d\theta  
 \end{equation}
 The penalty bears familiar derivative operations as  gradient descent, and therefore does not 
alter the algorithm's  complexity class. The strength of the penalty, $\lambda$, is akin to a continuously-variable Forget gate of an LSTM. Setting it to zero means that gradients for a batch are computed independent of any previous batch, while the deviation between the distribution of the two gradients is increasingly penalized with a larger $\lambda$. We present the calibration of $\lambda$ in Fig. $1$ of the appendix.

\section{Experiments}

To compare the effectiveness of $C^{3}$  we used 40 real-world benchmarking datasets. We use 13 image-based datasets benchmarks and 27 binary datasets from {KEEL} repository \cite{alcala2011keel} to evaluate our method. 
\begin{table}[h]
\tiny
\centering
\begin{tabular}{|l|c|c|c|c|c|c|c|c|} 
\hline
\multirow{2}{*}{Dataset} & Baseline & \multicolumn{4}{c|}{~ ~ ~ ~ ~ ~ ~ ~ ~SOTA} & Ours  & \multicolumn{2}{c|}{\begin{tabular}[c]{@{}c@{}}$\Delta_1$ = $C^3$ - CV\\$\Delta_2$ = $C^3$ - DIW\end{tabular}}  \\ 
\cline{2-9}
                         & CV       & IW   & IWCV & KMM  & DIW                   & $C^3$ & $\Delta_1(\%)$  & $\Delta_2(\%)$                                                                                \\ 
\cline{1-9}
MNIST                    & 94.8     & 85.1 & 76.9 & 11.8 & 98.0                  & 97.9  & $\uparrow$ 3.1  & $\downarrow$ 0.1                                                                              \\ 
\hline
Permuted-MNIST           & 95.1     & 67.8 & 75.4 & 11.2 & 84.7                  & 97.6  & $\uparrow$ 2.5  & $\uparrow$ 12.9                                                                               \\ 
\hline
Fashion-MNIST            & 82.3     & 80.2 & 72.3 & 10.3 & 87.2                  & 88.4  & $\uparrow$ 6.1  & $\uparrow$ 1.2                                                                                \\ 
\hline
Kuzushiji-MNIST          & 77.1     & 78.3 & 74.2 & 10.2 & 86.7                  & 89.2  & $\uparrow$ 12.1 & $\uparrow$ 2.5                                                                                \\ 
\hline
CIFAR-10                 & 71.5     & 79.8 & 69.9 & 9.61 & 80.4                  & 87.7  & $\uparrow$ 16.2 & $\uparrow$ 7.3                                                                                \\ 
\hline
CIFAR-100                & 38.2     & 44.6 & 51.7 & 8.53 & 53.6                  & 58.7  & $\uparrow$ 14.1 & $\uparrow$ 5.1                                                                                \\ 
\hline
CIFAR10-C                & 63.9     & 69.1 & 60.1 & 7.87 & 69.4                  & 73.3  & $\uparrow$ 9.40  & $\uparrow$ 3.9                                                                                \\ 
\hline
CIFAR100-C               & 28.8     & 18.7 & 16.9 & 5.37 & 32.6                  & 39.4  & $\uparrow$ 10.6 & $\uparrow$ 7.2                                                                                \\
\hline
\end{tabular}
\caption{$C^3$ vs SOTA.}
\label{tab: meanacc}
\end{table}\vspace{-2mm}

\section{Conclusions}

\begin{enumerate}
    \item Correcting causal covariate shift through $C^{3}$ also helps in natural covariate shift correction. $C^{3}$'s accuracy improves for a complete in-memory dataset with natural covariate shift such as Kuzushiji-MNIST, CIFAR10-C, CIFAR100-C, and Permuted-MNIST.
    \item We report increases of 16.2\%, 14.2\%, 12.1\%, 6.1\%, 2.5\% and 3.1\% accuracy as compares to standard CV (cross-validation) in a batchwise setup.
    \item We outperform SOTA benchmarks with improvements of 12.9\%, 7.3\%, and 5.1\% accuracy when we have access to the complete dataset. 
    \item $C^{3}$ also outperforms by 9.4\%, 7.7\%, and 7.4\% accuracy  improvement in k-fold setting CV.    
\end{enumerate}
Empirical results across several experimental baselines present evidence of our method serving as a short tether between varying covariate distributions, whenever data are distributed across batches, or experimentally as folds. The work therefore may be of immediate utility for federated or continual learning, or in AutoML. 

\subsubsection*{URM Statement}
We acknowledge that all authors of this work meet the URM criteria of ICLR 2024 Tiny Papers
Track.
\bibliography{c3iclr}
\bibliographystyle{iclr2023_conference_tinypaper}

\newpage
\appendix
\section{appendix}
\label{apndx}
In this section, we describe the relationship between relative entropy and fisher information. We also present the baselines, datasets details,  $ C^3$ batchwise performance $\lambda$ selection details, and experimental setup.
\subsection{Representing the current derivative with the Fisher
information matrix}
Let us consider having a model with parameter $\theta$ and a likelihood function \(p(X \mid \theta)\), where  \(X\) is observed data. The estimate of true parameter  $\theta$ can be found by using estimator $\hat{\theta}$. The Fisher information \(I(\theta)\) can be defined as the expected value of the negative hessian of the log-likelihood function.
\begin{equation}
I(\theta) = \mathbb{E}\left[- \frac{\partial^2 \log p(X\mid\theta)}{\partial\theta\partial\theta^T}\right]
\end{equation}
The Cram\'er-Rao Lower Bound (CRLB) states that for any unbiased estimator $\hat{\theta}$, the variance-covariance matrix \(V(\hat{\theta})\) satisfies the inequality property: 
\begin{equation}
 V(\hat{\theta}) \succeq I^{-1}(\theta) 
\end{equation}
The symbol $\succeq$ represents the following matrix inequality\(V(\hat{\theta}) - I^{-1}(\theta) \)  positive and semi-definite.

Now let us assume \(q(\hat{\theta}\)) as Gaussian distribution function to be   estimated around parameter $\theta$ mean and variance-covriance matrix \(V(\hat{\theta})\) in such manner that:
\begin{equation}
    q(\hat{\theta}) \approx \mathcal{N}(\theta, V(\hat{\theta}))  
\end{equation}
  We have a model $f(x)$ which outputs a target distribution $Q(y|x)$ for each given input $x$. The $D_{KL}$ divergence for source to target distribution can be found by $D_{KL}(P \,||\, Q) = \sum_y P(y|x) \log \left( \frac{P(y|x)}{Q(y|x)} \right)$.\\
  Considering y as continuous target variable, $P(y|x)$ and  $Q(y|x)$ as Gaussian distributions with means $\mu_{P}$ and $\mu_{Q}$ and variances $\sigma^2_{P}$ and $\sigma^2_{Q}$. \\
  Relative entropy can be computed in closed form using mean-variance of source and target distribution as follows: 
  \begin{equation}      
D_{KL}(P \,||\, Q) = \frac{1}{2} \left[ \log \left( \frac{\sigma^2_Q}{\sigma^2_P} \right) + \frac{\sigma^2_P + (\mu_P - \mu_Q)^2}{\sigma^2_Q} - 1 \right]
  \end{equation}
The $ D_{KL}$ can be approximated as:
\begin{equation}
  D_{KL}(p(\theta) \parallel q(\hat{\theta})) \approx \int p(\theta) \log\left( \frac{p(\theta)}{\mathcal{N}(\theta, V(\hat{\theta}))} \right) d\theta   
\end{equation}
With the help of CRLB  we can replace \(V(\hat{\theta})\) with \(I^{-1}(\theta)\) as \(V(\hat{\theta}) \succeq I^{-1}(\theta)\), we get:
\begin{equation}
 D_{KL}(p(\theta) \parallel q(\hat{\theta})) \approx \int p(\theta) \log\left( \frac{p(\theta)} {\mathcal{N}(\theta, I^{-1}(\theta))}\right) d\theta    
\end{equation}
which is the estimation of relative entropy by using a variance-covariance matrix of estimated parameters with the help of FIM.\\
\subsection{The Fisher information matrix as an approximation of variational posteriors}
Before introducing the penalty term which is one of our contributions we investigated the relation between relative entropy ($D_{KL}$) and  FIM.
Lets assume that $\theta$ is estimated parameter for given input data folds i.e ($X_1$,$X_2$,$X_3$,\ldots, $X_n$) with a probability function \(P(x;\theta)\). By using an unbiased estimator 
 \(\hat{\theta}(X_1, X_2, \ldots, X_n)\) of $\theta$, the variance estimator satisfies the following CRLB property. 
 \begin{equation}
   \text{$\sigma^2$}(\hat{\theta}) \geq \frac{1}{nI(\theta)}    
 \end{equation}

    where  \(I(\theta)\) is Fisher information and  \(n\) is sample size which can be described as:
\begin{equation}
 I(\theta) = -\mathbb{E}\left[\frac{\partial^2 \log P(X;\theta)}{\partial \theta^2}\right]  
\end{equation} 
It is crucial to understand the relation of  FIM to $D_{KL}$ ($D_{KL}$). We can find $D_{KL}$ between source to target  distributions \(P(x)\) and \(Q(x)\) with the same support set of X with K number of folds by:  
\begin{equation}
 D_{KL}(P\|Q) = \int_X P(x)\log\left(\frac{Q(x)}{P(x)}\right) dx     
\end{equation}
If we assume \(P(x;\theta)\) as true distribution for given input \(X\)  with parameter $\theta$ and \(Q(x;\hat{\theta}\)) as arbitrary target distribution with parameter $\hat{\theta}$  then we can rewrite $D_{KL}$ as:
\begin{equation}
  D_{KL}(P(\cdot;\theta) \parallel Q(\cdot;\hat{\theta})) = \mathbb{E}_{X \sim P(\cdot;\theta)}\left[\log\left(\frac{Q(X;\hat{\theta})}{P(X;\theta)}\right)\right]   
\end{equation}
Consider the special case of \(Q(x;\hat{\theta}\)) parameterized by $\hat\theta$ whereby we want to minimize  $D_{KL}$ w.r.t $\hat\theta$. For this case $D_{KL}$ is at minimum if we have \(Q(x;\hat{\theta}\)) = \(P(x;\hat{\theta}\)). Thus we get:
\begin{equation}
   D(P(\cdot;\theta) \parallel P(\cdot;\hat{\theta})) \geq 0  
\end{equation}
By applying Taylor expansion up to second-order to the log \(P(x;\hat{\theta}\)) for true parameter $\theta$ we have:
\begin{align}
\begin{split}
\log P(X;\hat{\theta}) &= \log P(X;\theta) \\
&\quad + (\hat{\theta} - \theta) \frac{\partial \log P(X;\theta)}{\partial \theta}
 \\
&\quad - \frac{1}{2} (\hat{\theta} - \theta)^2 \frac{\partial^2 \log P(X;\theta)}{\partial \theta^2} + O((\hat{\theta} - \theta)^3)
\end{split}
\end{align}
By taking expectation w.r.t \(X\)  we have:
\begin{align}
\begin{split}
\mathbb{E}_{X \sim P(\cdot;\theta)}\left[\log P(X;\hat{\theta}) - \log P(X;\theta)\right] &=\\ 
(\hat{\theta} - \theta)\mathbb{E}_{X \sim P(\cdot;\theta)}\left[\frac{\partial \log P(X;\theta)}{\partial \theta}\right]&\quad \\
- \frac{1}{2} (\hat{\theta} - \theta)^2 I(\theta) + O((\hat{\theta} - \theta)^3)
\end{split}
\end{align}
The left-hand in above mentioned equation is $D_{KL}$ i.e $D(P(\cdot;\theta) \parallel P(\cdot;\hat{\theta}))$. As we know that $D_{KL}$ is always non-negative, as so the right-hand side must also be non-negative. Thus we get: 
\begin{equation}
 \left(\frac{\hat{\theta} - \theta}{2}\right)I(\theta) \geq 0   
\end{equation}
It will hold for any $\hat{\theta}$, from this we can conclude that: \begin{equation}
 I(\theta) \geq 0   
\end{equation} which is Fisher information.
The following algorithm \ref{alg:c3} provides an overview of our proposed method $C^3$. 

\begin{algorithm}[h]
  \caption{Dataset fragmentation and causal covariate shift correction}
  \label{alg:c3}
  \begin{algorithmic}
    \State \textbf{Require: } model \( f(\theta) \) parameterized by \( \theta \);
    \State \hspace*{2em} training dataset \( \mathcal{D}_{tr} \);
    \State \hspace*{2em} validation data \( \mathcal{D}_{v} \);
    \State \hspace*{2em} number of batches \( K \);
    \State \hspace*{2em} number of epochs \( T \);
  \end{algorithmic}
  
  \begin{algorithmic}[1]
     \Procedure{ShiftCorrection}{$\mathcal{D}_{tr}$, $\mathcal{D}_{v}$}
     \State Split \( \mathcal{D}_{tr} \) into \( K \) batches  
     \State Initialize \( f(\theta) \) and \( \mathcal{L}(x, y; \theta) \)  
      \For{\( \text{epoch} \gets 1 \) \textbf{to} \( T \)}
        \For{\( i \gets 1 \) \textbf{to} \( K \)}
          \For{\( j \gets i+1 \) \textbf{to} \( K \)}
            \State Compute \( D_{\text{KL}}(D_i, D_j) \)
            \State \textbf{for each pair} \( (D_i, D_j) \):
            \State \( \mathcal{L}(x, y; \theta) = \)
            \State \hspace{1em} \( - \int P(y(x)) \log(P(y|x; \theta)) \, d\theta \)
            \State \hspace{1em} \( - \lambda \times \int \frac{\partial^2 \log p(X\mid\theta)}{\partial\theta\partial\theta^T} d\theta \)
          \EndFor
        \EndFor
        \State Update \( f(\theta) \) using \( \mathcal{L}(x, y; \theta) \)
      \EndFor
      \State \textbf{return} \( f(\theta) \)
    \EndProcedure
  \end{algorithmic}
\end{algorithm}

\section{Experiments}
In this section, we demonstrate the efficacy of $C^{3}$ against multiple baseline settings for causal covariate shift and on the benchmarks for natural covariate shift as a surrogate.
\begin{enumerate}
\item \textbf{Baselines:}
   
There are five baselines in our experiment:
\begin{itemize}
    \item \textbf{B1:} \textit{Clean} Verifying the effectiveness of $C^{3}$ on datasets without any covariate shift. 
    \item  \textbf{B2:} \textit{Natural shift Consequences} Analyzing the performance of the $C^{3}$ in the presence of natural covariate shift.
    \item  \textbf{B3:} \textit{Causal shift Consequences} Analyzing $C^{3}$ performance in the presence of causal shift caused by dataset fragmentation.
    \item  \textbf{B4:} \textit{Loss Recaliberation} Recaliberating the loss function and then measure the performance of $C^{3}$.
    \item  \textbf{B5:} \textit{Correction} correction of natural covariate shift  via proxy with $C^{3}$.
\end{itemize}
    \item \textbf{Model architecture:} We used a five-layer convolutional neural network (CNN) with softmax cross-entropy loss. Our CNN model consists of 2 convolutional layers with pooling, and 3 fully connected layers. The model architecture for all image-based benchmarks remains consistent, for tabular datsets the model architecture differs from image-based but remains the same for all tabular datasets. We used a multi-layer perceptron network for tabular data with a hidden layer with 4 neurons, relu as an activation function, and Adam optimizer. We set the hyper-parameter $\lambda$ value within the range (0.01, 0.04, 0.07, 0.1) in all of our experiments. We present $\lambda$ = 0.1 results in this paper for all of our experiments.   
All of our baselines are implemented in TensorFlow 2.11 \footnote{www.tensorflow.org} and the code is anonymously available at  \footnote{https://anonymous.4open.science/r/C3-C908/MNIST-Batchwise}.
\item \textbf{Machine Specification:} We run all of our experiments on RTX 3090 Ti with 24 GB GPU memory and 128 GB system memory.
\item \textbf{Benchmarks:} We compare the performance of $C^3$  with standard cross validation and significant importance based methods like: importance weighting (IW) \cite{huang2006correcting}, importance weighting cross-validation (IWCV) \cite{sugiyama2007covariate}, kernel mean matching (KMM) \cite{gretton2009covariate} and dynamic importance weighting (DIW) \cite{fang2020rethinking}. They were strategically chosen to represent landmark literature and current state-of-the-art.
\item \textbf{Datasets:} 
To compare the effectiveness of our developed method $C^{3}$ we used 40 real-world benchmarking datasets. To evaluate our method we used 13 image-based datasets benchmarks and 27 binary datasets from {KEEL} repository as benchmarks \cite{alcala2011keel}. The used image-based benchmarks, comprising: {MNIST} \cite{lecun1998mnist}, {Fashion-MNIST} \cite{xiao2017fashion}, {Kuzushiji-MNIST} \cite{clanuwat2018deep}, {Permuted-MNIST} \cite{goodfellow2013empirical}, {MNIST-C}  \cite{mu2019mnist} , {SVHN}  \cite{Netzer2011}, {Caltech101}  \cite{FeiFei2004LearningGV}, {Tiny ImageNet}  \cite{krizhevsky2009learning}, {STL-10} \cite{coates2011analysis} {CIFAR-10 and CIFAR-100}  \cite{krizhevsky2009learning}, {CIFAR10-C and CIFAR100-C}  \cite{hendrycks2019robustness}.
\item \textbf{Calibrating the penalty:} One of our contributions to the paper is introducing the penalty term as above mentioned. We calibrate the penalty term ($\lambda$) with different values in batch/fold setup and present the result in Fig \ref{fig:lambda}. We report $\lambda = 0.1$ results in our paper because our method $C^3$ is more robust to covariate shift. In figure \ref{fig:lambda} we can observe that $C^3$ performs better when we set $\lambda = 0.1$ as compared to other values.  
\begin{figure}[H]
    \centering
    \includegraphics[width=0.5\textwidth]{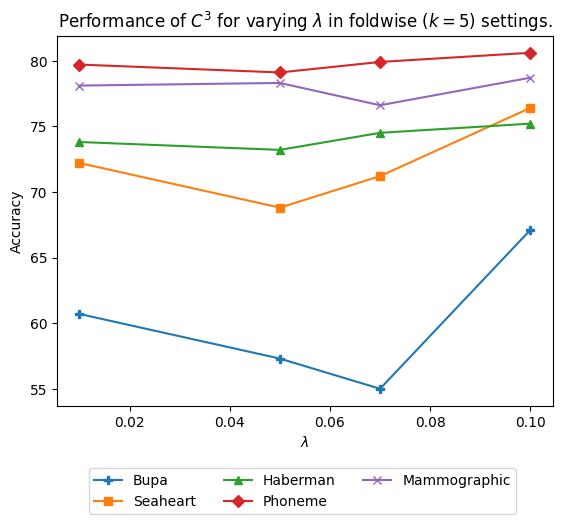}
    \caption{ Performance of $C^3$ for varying $\lambda$ in foldwise ($k = 5$) settings.}
    \label{fig:lambda}
\end{figure}

\item  \textbf{Experimental design:} To study the effect of causal covariate shift caused by fragmentation, we perform evaluations on datasets with natural covariate shift and also on clean (free of covariate shift) datasets. We use accuracy as the first and direct evaluation metric in all experiments. We run each experiment 5 times and report average results due to spatial constraints. 
\end{enumerate}
\section{Discussion}
To verify the effectiveness of $C^{3}$, we perform batchwise experiments for causal covariate shift whose results are presented in Table \ref{tab: ccca} which also validates \textbf{B4 \& B5}. We consider batchwise holdout cross-validation as a baseline in comparison to $C^{3}$. To ensure better performance of $C^{3}$ we compare the mean accuracy over all batches $\mu_1$ of Table \ref{tab: bcv} and $\mu_2$ of Table \ref{tab: ccca}. We report accuracy for each single batch as well in all experimental settings to verify $C^{3}$ performance. We then consider $C^{3}$ with the whole dataset as a baseline for our $C^{3}$ batchwise method.\\
The $\Delta_{3}$ of Table \ref{tab: ccca} presents the difference between $\mu_2$ of Table \ref{tab: ccca} and $\mu_1$ of Table \ref{tab: bcv}. The $\Delta_{3}$ shows improvement in accuracy and provides support to  our claim of causal covariate shift correction \textbf{B5}.
To verify \textbf{B5} we executed $C^{3}$ in batchwise settings on all dataset which results are reported in Table \ref{tab: ccca}.\\

Our proposed method $C^{3}$, shows improvement in accuracy in almost every batchwise setting and for each batch also as compared to the baseline.
To validate the adaptive nature of $C^{3}$ to natural shift correction, we perform experiments on above mentioned datasets with natural shift. We notice that $C^{3}$ is able to correct natural shift when it tries to correct causal shift. $C^{3}$ shows improvement in accuracy for almost all benchmarks, like it shows 5\%, 13.9\%, and 8.6\% improvement for Kuzushiji-MNIST, CIFAR10-C, and Fashion-MNIST with 20 batch split. 
$C^{3}$ also adapts to natural shift when it tries to correct causal covariate shift. \\
$C^{3}$'s accuracy improves as the number of batches decreases, due to statistics getting more robust with larger supports. It is shown in Table \ref{tab: ccca}. $C^{3}$  7.5\% for Fashion-MNIST and 6.9\% improvement in accuracy in 10 batch setup as compared to CV with the same batch setup. 
$C^{3}$ improves in accuracy with 7.2\%, 7.2\%, and 2.1\%  for Fashion-MNIST, Kuzushiji-MNIST, and Permuted-MNIST when batch size is 6. For the batch size 5, the improvement is 9.7\%,8.1\%, and 7.3\% for CIFAR100, Kuzushiji-MNIST, and Fashion-MNIST. In 4 batches scenario we report 11.3\%, 6.9\%, 6.1\%, and 5.8\% improvement in accuracy for CIFAR-100, Khushiji-MNIST, Fashion-MNIST, and CIFAR100-C. In the case of 2 batches the improvement in accuracy is 20.3\%, 15.5\%, 6.6\%, and 6.3\% for CIFAR-10, CIFAR-10, CIFAR100-C, and Khushiji-MNIST. Overall, $C^{3}$ outperforms in the batchwise case and in the case where a complete dataset is provided with other benchmarking methods, and results are discussed ahead in the comparison with SOTA.
\FloatBarrier
\begin{table}[h]
\tiny
\centering
\begin{tabular}{|l|l|l|l|l|l|l|l|l|c|l|c|} 
\hline
\multirow{2}{*}{\textbf{Dataset}} & \multicolumn{2}{c|}{Baseline} & \multicolumn{6}{c|}{Batchwise accuracy}                                                   & Mean                      & Variance     & $\Delta_3 = \mu_2 - \mu_1$                        \\ 
\cline{2-12}
                                  & \textbf{CV} & $\textbf{C}^3$  & \textbf{B1} & \textbf{B2} & \textbf{B3}   & \textbf{B4}   & \textbf{B5}   & \textbf{B6}   & $\mu_2$                   & $\sigma_2^2$ & $\Delta_3(\%)$                                    \\
   \hline                               
\multicolumn{12}{|c|}{\textbf{Training data = 5\% , Number\_of\_Batches = 20}}                                                                                                                                                                               \\ 
\hline
MNIST                             & 94.8        & 97.9            & 90.7        & 90.6        & 91            & 91.7          & 91.4          & 91.8          & 91.2                      & 0.09         & $\uparrow 2.5$                                    \\ 
\hline
Permuted-MNIST                    & 95.1        & 97.6            & 91.1        & 89.8        & 90.2          & 90.4          & 91.5          & 90.7          & 90.3                      & 0.26         & $\uparrow 1.9$                                    \\ 
\hline
Fashion-MNIST                     & 83.1        & 88.4            & 81.5        & 81.7        & 81.2          & 81.4          & 81.5          & 81.9          & 81.5                      & 0.058        & $\uparrow 8.6$                                    \\ 
\hline
Kuzushiji-MNIST                   & 75.4        & 89.2            & 68.5        & 69.4        & 67.2          & 68            & 67.8          & 66.9          & 68.4                      & 0.63         & $\uparrow 5.0$                                    \\ 
\hline
CIFAR-10                          & 71.5        & 88.7            & 50.9        & 51.4        & 52.2          & 48.9          & 50.3          & 57.4          & 51.8                      & 7.18         & $\uparrow 1.9$                                    \\ 
\hline
CIFAR-100                         & 38.2        & 58.7            & 23.9        & 18.2        & 18.5          & 17.8          & 23.9          & 18.3          & 20.1                      & 7.26         & $\uparrow 1.8$                                    \\ 
\hline
CIFAR10-C                         & 63.9        & 73.3            & 46.4        & 54.3        & 57.8          & 61.1          & 61.5          & 61.8          & \multicolumn{1}{l|}{57.2} & 30.1         & \multicolumn{1}{l|}{~ ~ ~ ~ ~ ~$\uparrow 13.9$}   \\ 
\hline
CIFAR100-C                        & 28.8        & 39.4            & 11.9        & 17.2        & 18.5          & 21.3          & 22.1          & 24.9          & \multicolumn{1}{l|}{19.3} & 17.1         & \multicolumn{1}{l|}{~ ~ ~ ~ ~ ~$\uparrow 0.9$}    \\ 
\hline
\multicolumn{12}{|c|}{\textbf{Training data = 10\% , Number\_of\_Batches = 10}}                                                                                                                                                                              \\ 
\hline
MNIST                             & 94.8        & 97.9            & 91.9        & 91.7        & 91.2          & 91.8          & 91.3          & 91.8          & 91.7                      & 0.08         & $\uparrow 0.6$                                    \\ 
\hline
Permuted-MNIST                    & 95.1        & 97.6            & 91.6        & 91.9        & 91.3          & 91.6          & 90.1          & 91.2          & 91.5                      & 0.31         & 0                                                 \\ 
\hline
Fashion-MNIST                     & 83.1        & 88.4            & 79.5        & 82.4        & 81.6          & 79.5          & 82.3          & 81.9          & 81.2                      & 1.21         & $\uparrow 7.5$                                    \\ 
\hline
Kuzushiji-MNIST                   & 75.4        & 89.2            & 71.4        & 70.4        & 71.7          & 70.7          & 70.5          & 70.9          & 70.9                      & 0.71         & $\uparrow 6.9$                                    \\ 
\hline
CIFAR-10                          & 71.5        & 88.7            & 52.1        & 53.1        & 48.5          & 59.3          & 52.5          & 55.7          & 53.5                      & 11.09        & $\uparrow 0.9$                                    \\ 
\hline
CIFAR-100                         & 38.2        & 58.7            & 27.2        & 25.8        & 20.4          & 17.0          & 21.9          & 22.8          & 22.5                      & 11.3         & $\uparrow 1.0$                                    \\ 
\hline
CIFAR10-C                         & 63.9        & 73.3            & 52.7        & 59.9        & 61.9          & 64.4          & 66.1          & 65.7          & \multicolumn{1}{l|}{61.7} & 21.1         & \multicolumn{1}{l|}{~ ~ ~ ~ ~ ~$\uparrow 39.1$}   \\ 
\hline
CIFAR100-C                        & 28.8        & 39.4            & 16.2        & 22.1        & 24.8          & 27.2          & 26.8          & 27.3          & \multicolumn{1}{l|}{21.1} & 15.6         & \multicolumn{1}{l|}{~ ~ ~ ~ ~ ~$\downarrow 1.7$}  \\ 
\hline
\multicolumn{12}{|c|}{\textbf{Training data = 15\% , Number\_of\_Batches = 6 }}                                                                                                                                                                              \\ 
\hline
MNIST                             & 94.8        & 97.9            & 93.2        & 92.6        & 93.2          & 93.5          & 93.1          & 93.3          & 93.2                      & 0.09         & $\uparrow 1.7$                                    \\ 
\hline
Permuted-MNIST                    & 95.1        & 97.6            & 93.1        & 93          & 92.7          & 93.7          & 93.5          & 93.1          & 93.2                      & 0.13         & $\uparrow 2.1$                                    \\ 
\hline
Fashion-MNIST                     & 83.1        & 88.4            & 81.4        & 81.9        & 82.9          & 80.9          & 82.3          & 82.2          & 81.9                      & 0.49         & $\uparrow 7.2$                                    \\ 
\hline
Kuzushiji-MNIST                   & 75.4        & 89.2            & 73.8        & 73.9        & 74.4          & 74.6          & 73.9          & 74.2          & 74.2                      & 0.13         & $\uparrow 7.2$                                    \\ 
\hline
CIFAR-10                          & 71.5        & 88.7            & 53.4        & 57.8        & 55.9          & 54.1          & 56.1          & 58.2          & 55.9                      & 3.7          & $\uparrow 2.0$                                    \\ 
\hline
CIFAR-100                         & 38.2        & 58.7            & 25.8        & 22.4        & 26.6          & 23.4          & 22.9          & 24.7          & 24.3                      & 2.34         & $\uparrow 1.7$                                    \\ 
\hline
CIFAR10-C                         & 63.9        & 73.3            & 58.4        & 61.1        & 63.2          & 64.3          & 67.4          & 66.9          & \multicolumn{1}{l|}{63.5} & 9.87         & \multicolumn{1}{l|}{~ ~ ~ ~ ~ ~$\uparrow 42.4$}   \\ 
\hline
CIFAR100-C                        & 28.8        & 39.4            & 21.9        & 25.5        & 27.2          & 28.7          & 28.5          & 29.5          & \multicolumn{1}{l|}{26.8} & 6.6          & \multicolumn{1}{l|}{~ ~ ~ ~ ~ ~$\uparrow 1.2$}    \\ 
\hline
\multicolumn{12}{|c|}{\textbf{Training data = 20\% , Number\_of\_Batches = 5 }}                                                                                                                                                                              \\ 
\hline
MNIST                             & 94.8        & 97.9            & 93.6        & 93.8        & 94.3          & 93.7          & 93.8          & $\textendash$ & 93.8                      & 0.07         & $\uparrow 1.9$                                    \\ 
\hline
Permuted-MNIST                    & 95.1        & 97.6            & 94.1        & 93.9        & 93.6          & 94.1          & 94.3          & $\textendash$ & 94                        & 0.07         & $\uparrow 2.2$                                    \\ 
\hline
Fashion-MNIST                     & 83.1        & 88.4            & 82.8        & 83.1        & 82.1          & 81.4          & 82.6          & $\textendash$ & 82.4                      & 0.44         & $\uparrow 7.3$                                    \\ 
\hline
Kuzushiji-MNIST                   & 75.4        & 89.2            & 75.4        & 76.3        & 75.8          & 75.1          & 75.6          & $\textendash$ & 75.6                      & 0.21         & $\uparrow 8.1$                                    \\ 
\hline
CIFAR-10                          & 71.5        & 88.7            & 50.3        & 56.2        & 53.5          & 57.8          & 59.9          & $\textendash$ & 55.5                      & 11.3         & $\uparrow 18$                                     \\ 
\hline
CIFAR-100                         & 38.2        & 58.7            & 34.2        & 35.4        & 33.9          & 34.9          & 34.7          & $\textendash$ & 34.6                      & 11.7         & $\uparrow 9.7$                                    \\ 
\hline
CIFAR10-C                         & 63.9        & 73.3            & 58.4        & 63.3        & 64.3          & 66.5          & 66.2          & $\textendash$ & \multicolumn{1}{l|}{63.7} & 8.53         & \multicolumn{1}{l|}{~ ~ ~ ~ ~ ~$\uparrow 1.7$}    \\ 
\hline
CIFAR100-C                        & 28.8        & 39.4            & 22.1        & 25.4        & 27.4          & 28.9          & 29.7          & $\textendash$ & \multicolumn{1}{l|}{26.7} & 7.43         & \multicolumn{1}{l|}{~ ~ ~ ~ ~ ~$\uparrow 2.9$}    \\ 
\hline
\multicolumn{12}{|c|}{\textbf{Training data = 25\% , Number\_of\_Batches = 4 }}                                                                                                                                                                              \\ 
\hline
MNIST                             & 94.8        & 97.9            & 94.4        & 94.4        & 94.3          & 94.4          & $\textendash$ & $\textendash$ & 94.4                      & 0.003        & $\uparrow 2.0$                                    \\ 
\hline
Permuted-MNIST                    & 95.1        & 97.6            & 94.4        & 94.3        & 94.5          & 94.5          & $\textendash$ & $\textendash$ & 94.4                      & 0.009        & $\uparrow 2.8$                                    \\ 
\hline
Fashion-MNIST                     & 83.1        & 88.4            & 82.8        & 83.2        & 83.6          & 83.5          & $\textendash$ & $\textendash$ & 83.3                      & 0.13         & $\uparrow 6.1$                                    \\ 
\hline
Kuzushiji-MNIST                   & 75.4        & 89.2            & 77.2        & 75.5        & 77.6          & 75.3          & $\textendash$ & $\textendash$ & 76.4                      & 1.37         & $\uparrow 6.9$                                    \\ 
\hline
CIFAR-10                          & 71.5        & 88.7            & 56.8        & 57.3        & 62.2          & 63.4          & $\textendash$ & $\textendash$ & 59.9                      & 8.47         & $\uparrow 5.0$                                    \\ 
\hline
CIFAR-100                         & 38.2        & 58.7            & 33.9        & 34.1        & 34.7          & 33.5          & $\textendash$ & $\textendash$ & 34.1                      & 0.18         & ~ ~$\uparrow 11.3$                                \\ 
\hline
CIFAR10-C                         & 63.9        & 73.3            & 60.9        & 64.4        & 66.8          & 68.3          & $\textendash$ & $\textendash$ & \multicolumn{1}{l|}{65.1} & 7.81         & \multicolumn{1}{l|}{~ ~ ~ ~ ~ ~$\uparrow 3.7$}    \\ 
\hline
CIFAR100-C                        & 28.8        & 39.4            & 24.7        & 28.2        & 29.7          & 31.9          & $\textendash$ & $\textendash$ & \multicolumn{1}{l|}{28.6} & 6.86         & \multicolumn{1}{l|}{~ ~ ~ ~ ~ ~$\uparrow 5.8$}    \\ 
\hline
\multicolumn{12}{|c|}{\textbf{Training data = 50\% , Number\_of\_Batches = 2 }}                                                                                                                                                                              \\ 
\hline
MNIST                             & 94.8        & 97.9            & 95.9        & 96.1        & $\textendash$ & $\textendash$ & $\textendash$ & $\textendash$ & 96                        & 0.02         & $\uparrow 2.5$                                    \\ 
\hline
Permuted-MNIST                    & 95.1        & 97.6            & 95.7        & 96.1        & $\textendash$ & $\textendash$ & $\textendash$ & $\textendash$ & 95.9                      & 0.08         & $\uparrow 2.4$                                    \\ 
\hline
Fashion-MNIST                     & 83.1        & 88.4            & 84.2        & 84.4        & $\textendash$ & $\textendash$ & $\textendash$ & $\textendash$ & 84.3                      & 0.02         & $\uparrow 4.5$                                    \\ 
\hline
Kuzushiji-MNIST                   & 75.4        & 89.2            & 79.3        & 80.4        & $\textendash$ & $\textendash$ & $\textendash$ & $\textendash$ & 79.8                      & 0.61         & $\uparrow 6.3$                                    \\ 
\hline
CIFAR-10                          & 71.5        & 88.7            & 76.3        & 80.6        & $\textendash$ & $\textendash$ & $\textendash$ & $\textendash$ & 78.4                      & 4.62         & ~ ~$\uparrow 20.3$                                \\ 
\hline
CIFAR-100                         & 38.2        & 58.7            & 39.8        & 39.9        & $\textendash$ & $\textendash$ & $\textendash$ & $\textendash$ & 39.85                     & .002         & ~ ~$\uparrow 15.5$                                \\ 
\hline
CIFAR10-C                         & 63.9        & 73.3            & 65.5        & 68.6        & $\textendash$ & $\textendash$ & $\textendash$ & $\textendash$ & \multicolumn{1}{l|}{67.1} & 2.4          & \multicolumn{1}{l|}{~ ~ ~ ~ ~ ~$\uparrow 2.8$}    \\ 
\hline
CIFAR100-C                        & 28.8        & 39.4            & 31.2        & 34.8        & $\textendash$ & $\textendash$ & $\textendash$ & $\textendash$ & \multicolumn{1}{l|}{33}   & 3.24         & \multicolumn{1}{l|}{~ ~ ~ ~ ~ ~$\uparrow 6.6$}    \\
\hline
\end{tabular}
\caption{$C^3$ Batchwise Accuracy }
\label{tab: ccca}
\end{table}
\FloatBarrier

\FloatBarrier
\begin{table}
\tiny
\centering
\begin{tabular}{|l|l|l|c|c|c|c|l|l|c|c|} 
\hline
\multirow{2}{*}{\textbf{Dataset }} & \multicolumn{2}{l|}{Baseline} & \multicolumn{6}{l|}{Batchwise accuracy}                                                                                                                                                               & Mean                      & Variance                            \\ 
\cline{2-11}
                                   & \textbf{CV} & \textbf{$C^3$}  & \multicolumn{1}{l|}{\textbf{B1}} & \multicolumn{1}{l|}{\textbf{B2}} & \multicolumn{1}{l|}{\textbf{B3}}   & \multicolumn{1}{l|}{\textbf{B4}}   & \textbf{B5}               & \textbf{B6}               & $\mu_1$                   & $\sigma_1^2$                        \\
                                   \hline
\multicolumn{11}{|l|}{~ ~ ~ ~ ~ ~ ~ ~ ~ ~ ~ ~ ~ ~ ~ ~ ~ ~ ~ ~ ~ ~\textbf{\textbf{Training data = 5\% , Number\_of\_Batches = 20 }}}                                                                                                                                                                                                          \\ 
\hline
MNIST                              & 94.8        & 97.9            & 88.1                             & 89.3                             & 87.9                               & 89.9                               & \multicolumn{1}{c|}{88.9} & \multicolumn{1}{c|}{88.8} & 88.7                      & 0.49                                \\ 
\hline
Permuted-MNIST                     & 95.1        & 97.6            & 88                               & 86.1                             & 88.9                               & 88.7                               & \multicolumn{1}{c|}{87.2} & \multicolumn{1}{c|}{88.3} & 88.4                      & 1.41                                \\ 
\hline
Fashion-MNIST                      & 83.1        & 88.4            & 72.7                             & 73.7                             & 74.2                               & 72.5                               & \multicolumn{1}{c|}{74}   & \multicolumn{1}{c|}{70.6} & 72.9                      & 1.94                                \\ 
\hline
Kuzushiji-MNIST                    & 75.4        & 89.2            & 63.6                             & 63.4                             & 62.2                               & 66.7                               & \multicolumn{1}{c|}{58.3} & \multicolumn{1}{c|}{63.4} & 63.4                      & 5.59                                \\ 
\hline
CIFAR-10                           & 71.5        & 88.7            & 44.1                             & 49.0                             & 50.3                               & 50.7                               & \multicolumn{1}{c|}{51.2} & \multicolumn{1}{c|}{54.5} & 49.9                      & 9.67                                \\ 
\hline
CIFAR-100                          & 38.2        & 58.7            & 13.1                             & 16.5                             & 18.1                               & 19.3                               & \multicolumn{1}{c|}{20.4} & \multicolumn{1}{c|}{22.7} & 18.3                      & 9.17                                \\ 
\hline
CIFAR10-C                          & 63.9        & 73.3            & \multicolumn{1}{l|}{19.3}        & \multicolumn{1}{l|}{20.1}        & \multicolumn{1}{l|}{16.3}          & \multicolumn{1}{l|}{16.1}          & 14.9                      & 10.2                      & \multicolumn{1}{l|}{16.2} & \multicolumn{1}{l|}{~ ~ 10.4}       \\ 
\hline
CIFAR100-C                         & 28.8        & 39.4            & \multicolumn{1}{l|}{11.1}        & \multicolumn{1}{l|}{16.3}        & \multicolumn{1}{l|}{19.1}          & \multicolumn{1}{l|}{19.7}          & 21.6                      & 22.3                      & \multicolumn{1}{l|}{18.4} & \multicolumn{1}{l|}{~~ ~14.2}           \\ 
\hline
\multicolumn{11}{|l|}{~ ~ ~ ~ ~ ~ ~ ~ ~ ~ ~ ~ ~ ~ ~ ~ ~ ~ ~ ~ ~\textbf{Training data = 10\% , Number\_of\_Batches = 10}}                                                                                                                                                                                                                     \\ 
\hline
MNIST                              & 94.8        & 97.9            & 90.7                             & 91.7                             & 91.1                               & 90.2                               & \multicolumn{1}{c|}{89.3} & \multicolumn{1}{c|}{91.5} & 91.1                      & 1.06                                \\ 
\hline
Permuted-MNIST                     & 95.1        & 97.6            & 91.9                             & 92.8                             & 91.8                               & 91.2                               & \multicolumn{1}{c|}{91.4} & \multicolumn{1}{c|}{91.3} & 91.5                      & 0.62                                \\ 
\hline
Fashion-MNIST                      & 83.1        & 88.4            & 69.4                             & 69.9                             & 75.2                               & 73.4                               & \multicolumn{1}{c|}{74.4} & \multicolumn{1}{c|}{71.7} & 73.7                      & 9.22                                \\ 
\hline
Kuzushiji-MNIST                    & 75.4        & 89.2            & 64                               & 63.6                             & 64.1                               & 64.9                               & \multicolumn{1}{c|}{65.2} & \multicolumn{1}{c|}{64.5} & 64.0                      & 1.15                                \\ 
\hline
CIFAR-10                           & 71.5        & 88.7            & 47.9                             & 52.8                             & 52.9                               & 53.7                               & \multicolumn{1}{c|}{54.5} & \multicolumn{1}{c|}{54.1} & 52.6                      & 4.87                                \\ 
\hline
CIFAR-100                          & 38.2        & 58.7            & 17.3                             & 20.3                             & 22.2                               & 23.3                               & \multicolumn{1}{c|}{24.7} & \multicolumn{1}{c|}{21.2} & 21.5                      & 5.51                                \\ 
\hline
CIFAR10- C                         & 63.9        & 73.3            & \multicolumn{1}{l|}{22.8}        & \multicolumn{1}{l|}{17.6}        & \multicolumn{1}{l|}{18.4}          & \multicolumn{1}{l|}{12.2}          & 17.4                      & 12.9                      & \multicolumn{1}{l|}{22.6} & \multicolumn{1}{l|}{~ ~  16.8}      \\ 
\hline
CIFAR100-C                         & 28.8        & 39.4            & \multicolumn{1}{l|}{15.5}        & \multicolumn{1}{l|}{21.9}        & \multicolumn{1}{l|}{25.2}          & \multicolumn{1}{l|}{26.6}          & 27.1                      & 20.5                      & \multicolumn{1}{l|}{22.8} & \multicolumn{1}{l|}{~ ~      16.3}  \\ 
\hline
\multicolumn{11}{|l|}{\textbf{ ~ ~ ~ ~ ~ ~ ~ ~ ~ ~ ~ ~ ~ ~ ~ ~ ~ ~ ~Training data = 15\% , Number\_of\_Batches = 6}}                                                                                                                                                                                                                         \\ 
\hline
MNIST                              & 94.8        & 97.9            & 90.8                             & 90.7                             & 91.7                               & 92.2                               & \multicolumn{1}{c|}{92.2} & \multicolumn{1}{c|}{91.6} & 91.5                      & 0.43                                \\ 
\hline
Permuted-MNIST                     & 95.1        & 97.6            & 91.2                             & 90.2                             & 91.2                               & 92.1                               & \multicolumn{1}{c|}{90.6} & \multicolumn{1}{c|}{91.1} & 91.1                      & 0.41                                \\ 
\hline
Fashion-MNIST                      & 83.1        & 88.4            & 74.4                             & 75.7                             & 77.7                               & 75                                 & \multicolumn{1}{c|}{70.6} & \multicolumn{1}{c|}{74.7} & 74.7                      & 5.40                                \\ 
\hline
Kuzushiji-MNIST                    & 75.4        & 89.2            & 66.9                             & 67                               & 66.4                               & 69.9                               & \multicolumn{1}{c|}{64.8} & \multicolumn{1}{c|}{67.2} & 67.0                      & 2.73                                \\ 
\hline
CIFAR-10                           & 71.5        & 88.7            & 50.9                             & 51.2                             & 53.9                               & 53.8                               & \multicolumn{1}{c|}{57.1} & \multicolumn{1}{c|}{56.9} & 53.9                      & 5.91                                \\ 
\hline
CIFAR-100                          & 38.2        & 58.7            & 18.2                             & 21.1                             & 22.5                               & 23.4                               & \multicolumn{1}{c|}{24.1} & \multicolumn{1}{c|}{24.4} & 22.6                      & 4.57                                \\ 
\hline
CIFAR10- C                         & 63.9        & 73.3            & \multicolumn{1}{l|}{18.3}        & \multicolumn{1}{l|}{25.1}        & \multicolumn{1}{l|}{23.9}          & \multicolumn{1}{l|}{20.7}          & 20.9                      & 21.1                      & \multicolumn{1}{l|}{21.6} & \multicolumn{1}{l|}{~ ~   4.99}     \\ 
\hline
CIFAR100- C                        & 28.8        & 39.4            & \multicolumn{1}{l|}{18.2}        & \multicolumn{1}{l|}{19.9}        & \multicolumn{1}{l|}{21.2}          & \multicolumn{1}{l|}{21.9}          & 25.6                      & 27.1                      & \multicolumn{1}{l|}{22.3} & \multicolumn{1}{l|}{~ ~   9.64}     \\ 
\hline
\multicolumn{11}{|l|}{~ ~ ~ ~ ~ ~ ~ ~ ~ ~ ~ ~ ~ ~ ~ ~ ~ ~ ~ ~ ~\textbf{Training data = 20\% , Number\_of\_Batches = 5}}                                                                                                                                                                                                                      \\ 
\hline
MNIST                              & 94.8        & 97.9            & 92.7                             & 92.2                             & 93.4                               & 91.2                               & \multicolumn{1}{c|}{90.2} & $\textendash $            & 91.9                      & 1.59                                \\ 
\hline
Permuted-MNIST                     & 95.1        & 97.6            & 91                               & 91.7                             & 91.4                               & 91.2                               & \multicolumn{1}{c|}{93.5} & $\textendash$             & 91.8                      & 1.01                                \\ 
\hline
Fashion-MNIST                      & 83.1        & 88.4            & 76.7                             & 74.1                             & 72                                 & 75.3                               & \multicolumn{1}{c|}{77.2} & $\textendash$             & 75.1                      & 4.40                                \\ 
\hline
Kuzushiji-MNIST                    & 75.4        & 89.2            & 68.3                             & 69.2                             & 67.7                               & 65.5                               & \multicolumn{1}{c|}{66.8} & $\textendash$             & 67.5                      & 2.02                                \\ 
\hline
CIFAR-10                           & 71.5        & 88.7            & 36.9                             & 38.5                             & 37.5                               & 36.8                               & \multicolumn{1}{c|}{37.9} & $\textendash$             & 37.5                      & 0.50                                \\ 
\hline
CIFAR-100                          & 38.2        & 58.7            & 19.7                             & 22.3                             & 23.5                               & 24.4                               & \multicolumn{1}{c|}{24.9} & $\textendash$             & 22.9                      & 3.43                                \\ 
\hline
CIFAR10- C                         & 63.9        & 73.3            & \multicolumn{1}{l|}{58.1}        & \multicolumn{1}{l|}{62.0}        & \multicolumn{1}{l|}{61.8}          & \multicolumn{1}{l|}{60.7}          & 67.5                      & $\textendash$             & \multicolumn{1}{l|}{62.0} & \multicolumn{1}{l|}{~ ~ 9.43}       \\ 
\hline
CIFAR100- C                        & 28.8        & 39.4            & \multicolumn{1}{l|}{20.0}        & \multicolumn{1}{l|}{23.2}        & \multicolumn{1}{l|}{23.2}          & \multicolumn{1}{l|}{26.7}          & 26.1                      & $\textendash$             & \multicolumn{1}{l|}{23.8} & \multicolumn{1}{l|}{~~ ~5.77}           \\ 
\hline
\multicolumn{11}{|l|}{~ ~ ~ ~ ~ ~ ~ ~ ~ ~ ~ ~ ~ ~ ~ ~ ~ ~ ~ ~ ~\textbf{Training data = 25\% , Number\_of\_Batches = 4}}                                                                                                                                                                                                                      \\ 
\hline
MNIST                              & 94.8        & 97.9            & 92.1                             & 93.6                             & 92.5                               & 91.3                               & $\textendash$             & $\textendash$             & 92.4                      & 0.92                                \\ 
\hline
Permuted-MNIST                     & 95.1        & 97.6            & 92.1                             & 92.2                             & 90.5                               & 91.6                               & $\textendash$             & $\textendash$             & 91.6                      & 0.61                                \\ 
\hline
Fashion-MNIST                      & 83.1        & 88.4            & 74.6                             & 77.5                             & 78.1                               & 78.5                               & $\textendash$             & $\textendash$             & 77.2                      & 3.12                                \\ 
\hline
Kuzushiji-MNIST                    & 75.4        & 89.2            & 70.6                             & 69.2                             & 69.5                               & 68.8                               & $\textendash$             & $\textendash$             & 69.5                      & 0.60                                \\ 
\hline
CIFAR-10                           & 71.5        & 88.7            & 51.9                             & 52.7                             & 56.9                               & 58.1                               & $\textendash$             & $\textendash$             & 54.9                      & 7.02                                \\ 
\hline
CIFAR-100                          & 38.2        & 58.7            & 20.3                             & 22.4                             & 23.9                               & 24.8                               & $\textendash$             & $\textendash$             & 22.8                      & 2.7                                 \\ 
\hline
CIFAR10- C                         & 63.9        & 73.3            & \multicolumn{1}{l|}{57.6}        & \multicolumn{1}{l|}{62.1}        & \multicolumn{1}{l|}{63.1}          & \multicolumn{1}{l|}{62.6}          & $\textendash$             & $\textendash$             & \multicolumn{1}{l|}{61.4} & \multicolumn{1}{l|}{~ ~ 4.81}       \\ 
\hline
CIFAR100- C                        & 28.8        & 39.4            & \multicolumn{1}{l|}{19.7}        & \multicolumn{1}{l|}{23.4}        & \multicolumn{1}{l|}{23.8}          & \multicolumn{1}{l|}{24.5}          & $\textendash$             & $\textendash$             & \multicolumn{1}{l|}{22.8} & \multicolumn{1}{l|}{~~ ~3.64}           \\ 
\hline
\multicolumn{11}{|l|}{~ ~ ~ ~ ~ ~ ~ ~ ~ ~ ~ ~ ~ ~ ~ ~ ~ ~ ~ ~ ~\textbf{Training data = 50\% , Number\_of\_Batches = 2}}                                                                                                                                                                                                                      \\ 
\hline
MNIST                              & 94.8        & 97.9            & 93.3                             & 93.7                             & \multicolumn{1}{l|}{$\textendash$} & \multicolumn{1}{l|}{$\textendash$} & $\textendash$             & $\textendash$             & 93.5                      & 0.08                                \\ 
\hline
Permuted-MNIST                     & 95.1        & 97.6            & 93.3                             & 93.7                             & \multicolumn{1}{l|}{$\textendash$} & \multicolumn{1}{l|}{$\textendash$} & $\textendash$             & $\textendash$             & 93.5                      & 0.08                                \\ 
\hline
Fashion-MNIST                      & 83.1        & 88.4            & 80                               & 79.7                             & \multicolumn{1}{l|}{$\textendash$} & \multicolumn{1}{l|}{$\textendash$} & $\textendash$             & $\textendash$             & 79.8                      & 0.04                                \\ 
\hline
Kuzushiji-MNIST                    & 75.4        & 89.2            & 73.6                             & 73.3                             & \multicolumn{1}{l|}{$\textendash$} & \multicolumn{1}{l|}{$\textendash$} & $\textendash$             & $\textendash$             & 73.5                      & 0.04                                \\ 
\hline
CIFAR-10                           & 71.5        & 88.7            & 56.2                             & 60.1                             & \multicolumn{1}{l|}{$\textendash$} & \multicolumn{1}{l|}{$\textendash$} & $\textendash$             & $\textendash$             & 58.1                      & 3.81                                \\ 
\hline
CIFAR-100                          & 38.2        & 58.7            & 23.2                             & 25.4                             & \multicolumn{1}{l|}{$\textendash$} & \multicolumn{1}{l|}{$\textendash$} & $\textendash$             & $\textendash$             & 24.3                      & 1.21                                \\ 
\hline
CIFAR10- C                         & 63.9        & 73.3            & \multicolumn{1}{l|}{63.2}        & \multicolumn{1}{l|}{65.5}        & \multicolumn{1}{l|}{$\textendash$} & \multicolumn{1}{l|}{$\textendash$} & $\textendash$             & $\textendash$             & \multicolumn{1}{l|}{64.3} & \multicolumn{1}{l|}{~~ ~1.32}           \\ 
\hline
CIFAR100- C                        & 28.8        & 39.4            & \multicolumn{1}{l|}{25.3}        & \multicolumn{1}{l|}{27.5}        & \multicolumn{1}{l|}{$\textendash$} & \multicolumn{1}{l|}{$\textendash$} & $\textendash$             & $\textendash$             & \multicolumn{1}{l|}{26.4} & \multicolumn{1}{l|}{~~ ~1.21}           \\
\hline
\end{tabular}
\caption{CV batch-wise accuracy}
\label{tab: bcv}
\end{table}
\FloatBarrier
\end{document}